# Annotation Graphs and Servers and Multi-Modal Resources: Infrastructure for Interdisciplinary Education, Research and Development

**Christopher Cieri**
University of Pennsylvania
Linguistic Data Consortium
3615 Market Street
Philadelphia, PA. 19104-2608 USA
`ccieri@ldc.upenn.edu`

**Steven Bird**
University of Pennsylvania
Linguistic Data Consortium
3615 Market Street
Philadelphia, PA. 19104-2608 USA
`sb@ldc.upenn.edu`

## Abstract

Annotation graphs and annotation servers offer infrastructure to support the analysis of human language resources in the form of time-series data such as text, audio and video. This paper outlines areas of common need among empirical linguists and computational linguists. After reviewing examples of data and tools used or under development for each of several areas, it proposes a common framework for future tool development, data annotation and resource sharing based upon annotation graphs and servers.

## 1 Introduction

Despite different methodologies, goals and traditions, researchers in a variety of specialties in linguistics and computational linguistics share a core of assumptions and needs. Research communities in empirical linguistics, natural language processing, speech recognition, information retrieval and language teaching have a common need for language resources such as observations of linguistic performance, annotations encoding human judgment, standards for maintaining consistency among distributed resources and processes for extracting relevant observations. Where needs overlap, there is the opportunity to reuse existing resources and coordinate new initiatives so that communities share the burden of development while benefiting from the results. Where computational linguistics interacts with other areas of language research and teaching, there are additional opportunities for symbiosis. Natural language technology may offer greater access and robustness to empirical linguistic research that in turn may offer new data necessary to develop new technologies. This paper discusses common infrastructure for the annotation of linguistic data and the application of that infrastructure to several traditionally very diverse fields of inquiry.

## 2 Common Assumptions, Needs and Goals in Natural Language Studies

Human language resources, expensive to create and maintain, are in increasing demand among a growing number of research communities. One solution to this expanding need is to reannotate and reuse language resources created for other purposes. The now classic example is that of the Switchboard-1 Corpus (ISBN: 1-58563-121-3), a collection of 2400 two-sided telephone conversations among 543 U.S. speakers, created by Texas Instruments in 1991. Although collected for speaker identification and topic spotting research, Switchboard has been widely used to support large vocabulary conversational speech recognition. It has been extensively corrected twice, once at Penn and NIST, and once at

Mississippi State. Two excerpts have been published as test corpora for government-sponsored projects. At least 6 other annotations have been created at various times and more-or-less widely distributed among research sites: part-of-speech annotation (Penn); syntactic structure annotation (Penn); dysfluency annotation (Penn); partial phonetic transcription (independently at UCLA and at Berkeley); and discourse function annotation (Colorado). These annotations use different "editions" of the underlying corpus and have sometimes silently introduced their own corrections or modified the data format to suit their needs. Thus the Colorado discourse function annotation was based on phrase structures introduced by the Penn dysfluency annotation, which in turn was based on the Penn/NIST corrections, which in turn were based on the original TI transcriptions of the underlying (and largely unchanging) audio files. Switchboard and its derivatives remain in active use worldwide, and new derivatives continue to be produced, along with (published and unpublished) corrections of old ones. This worsens the already acute problem of establishing and maintaining coherent relations among the derivatives in common use today.

The Switchboard-1 case is by no means isolated (Graff & Bird 2000). The Topic Detection and Tracking Corpus, TDT-2 (ISBN: 1-58563-157-4) was created in 1998 by LDC and contains newswire and more than 600 hours of transcribed broadcast news from 8 English and 3 Chinese sources sampled daily over six months with annotations to indicate story boundaries and relevance of those stories to 100 randomly selected topics. Since its release, TDT-2 has been used as training, development-test and evaluation data in the TDT evaluations; the audio has been used in TREC SDR evaluations (Garofalo, Auzanne and Voorhees 2000), TDT text has been partially re-annotated for entity detection in the Automatic Content Extraction project (Przybocki 2000) and portions have been used for the Center for Spoken Language Processing's workshops in Novel Information Detection (Allan et. al. 1999), Mandarin-English Information (Meng et. al. 2000) and Audio-Visual Speech Recognition (Chalapati 2000).

Switchboard and TDT are just two examples of a growing trend toward reannotation and reuse of language resources, a trend that is not limited to language engineering. Miller and Walker (2001) have demonstrated the value of the CALLHOME German corpus (ISBN: 1-58563-117-5), developed to support speech recognition research, for language teaching. Deckert & Yaeger-Dror (2000) have used Switchboard to study regional syntactic variation in American English.

Reannotation and reuse of linguistic data highlight the need for common infrastructure to support resource development across disciplines and specialties.

## 3  Overlaps between Human Language Technology and Other Linguistic Research

Many specialties in empirical linguistics and language engineering require large volumes of language data and tools for browsing and searching the data efficiently. The sections that follow provide examples of recent efforts to address emerging needs for language resources.

**Interlinear Texts and Linguistic Exploration**

Interlinear text is a product of linguistic fieldwork often in low-density languages. The physical appearance of interlinear text typically consists of a main text line annotated with linguistic transcriptions and analyses, such as morphological representations, glosses at various levels, part-of-speech tags, and a free translation at the sentence level. Fragments of these annotation lines are vertically aligned with the corresponding fragments of text. Phrasal translations and footnotes are often presented on other lines. Interlinear texts come in many forms and can be represented digitally in many ways, e.g. plain text with hard spacing, tables, special markup, and special-purpose data structures. There are various methods for linking to audio data and lexical entries, and for including footnotes and other marginalia. This diversity of form presents problems for general-purpose software for searching, exchanging, displaying and enriching interlinear texts. Nonetheless interlinear text is a precious resource with multiple uses in natural language processing. Its various components can be used in the development of lexical and morphological resources, can support tagging and parsing and can provide training material for machine translation. Maeda and Bird (2000, 2001) demonstrated a tool for creating interlinear text. A screenshot appears in Figure 1.

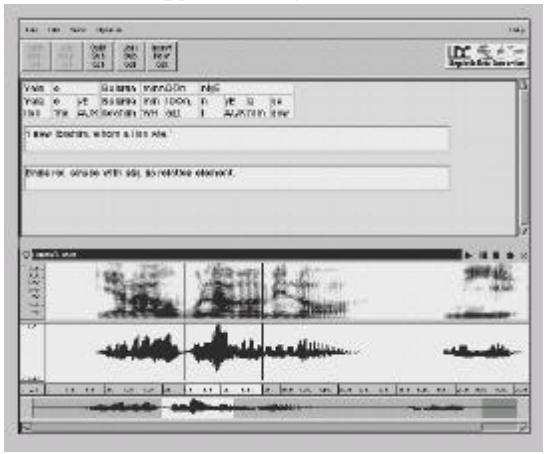

**Figure 1: Interlinear text tool using the AG Toolkit**

## Sociolinguistic Annotation

The quantitative analysis of linguistic variation begins with empirical observation and statistical description of linguistic behavior. Although general computer technology encourages the collection, annotation, analysis and discussion of linguistic behavior wholly within the digital domain, few tools exist to help the sociolinguist in this effort. The project on Data and Annotations for Sociolinguistics (DASL) is investigating best practices via a case study of well-documented sociolinguistic phenomena in several large speech corpora: TIMIT, Switchboard-1, CallHome and Hub-4. Researchers are currently annotating the corpora for t/d deletion, the process by which [t] and [d] sometimes fail to be realized under certain phonological, morphological and social conditions. The case study is also a means to address broader questions: How do the specified corpora compare with the interview data typically used in sociolinguistics? Will the study of corpus data reveal new patterns not evident in the more common studies conducted within the framework of the speech community? Can empirical research on language variation be organized on a large scale with teams of non-specialist annotators?

All of the data used in DASL were originally created to support human language technology development; the datasets are currently being reannotated to support empirical studies of linguistic variation. A custom annotation tool allows users to query each corpus for tokens of potential interest greatly reducing effort relative to traditional approaches. Annotators can read or listen to each token, access demographic data and encode their observations in formats compatible with other analytical software used in the community. The web-based interface in Figure 2 promotes multi-site annotation and the study of inter-annotator consistency (Cieri and Strassel, 2001).

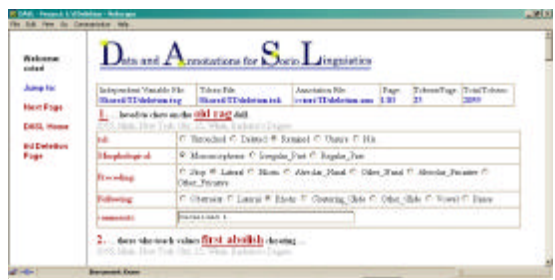

**Figure 2: Sociolinguistic Annotation Tool**

## Authoring Resources and Tools for Language Learning

Although current information technology encourages new approaches in computer assisted language learning and teaching, progress in this area is hampered by an inadequate supply of language resources. The SMART (Source Media Authoring Resources and Tools) pilot project is addressing this problem by providing appropriately licensed data and software resources for preparing language-learning material. The Linguistic Data Consortium, a partner in this effort, is contributing several of its large data sets including conversational and broadcast data in Arabic, English, French and German. The language resources overlap almost completely with those used in language engineering. SMART is building upon the distribution model established in LDC Online, a service that provides network-based access to hundreds of gigabytes of text and audio data and annotations. Audio data are available digitally in files corresponding to a conversation, broadcast or other linguistic event. To facilitate searching, LDC Online includes, according to their availability, human- and machine-generated transcripts time-aligned to permit more fine-grained access. For example, where a time-aligned transcript of a conversation exists, users may extract, reformat and play any segment specified by the time stamps in the transcript. SMART is building upon this foundation by providing additional data resources, browsing and search customized to the needs of language teachers and additional output formats to accommodate courseware authoring tools available in the commercial market.

SMART promises to benefit a wide range of language teachers and learners but only to the extent that its resources are readily available. The volume of SMART data exceeds that which can be easily transferred over a network. Even small video clips consume hundreds of megabits of bandwidth. Instead SMART data will be delivered via servers that maintain raw data and associated annotations, permit browsing and queries and allow the user to specify the format and granularity of the response. The user will have the option of downloading the data for local use or adding annotations that may be kept privately or made public via the annotation server. The technology of the annotation server coupled with the extensibility of annotation graphs described below will enables nearly unconstrained access to SMART data.

These efforts to support interlinear text, sociolinguistic annotation and multimodal data in language teaching each require flexible access to signal data and associated annotations. The sections that follow describe an architecture that provides such access.

## 4 Annotation Graphs, Annotation Servers and a Query Language: Common Infrastructure for Coordinated Research, Resource Development

Storing and serving large amounts of annotated data via the web requires interoperable data representations and tools along with methods for handling external formats and protocols for querying and delivering annotations. Annotation graphs were presented by Bird and Liberman (1999) as a general purpose model for representing and manipulating annotations of time series data, regardless of their physical storage format. An annotation graph is a labeled, directed, acyclic graph with time offsets on some of its nodes. The formalism is illustrated below by application to the TIMIT Corpus (Garofalo et al, 1986). The original TIMIT word file contains starting and ending offsets (in 16KHz samples) and transcripts of each word in the audio file

```
train/dr1/fjsp0/sa1.wrd:
  2360    5200    she
  5200    9680    had
  9680   11077    your
 11077   16626    dark
 16626   22179    suit
 22179   24400    in
 24400   30161    greasy
 30161   36150    wash
 36720   41839    water
 41839   44680    all
 44680   49066    year
```

The phone file provides the same information for each sound in the audio file. This is the phonetic transcription for "she had".

```
train/dr1/fjsp0/sa1.phn:
     0    2360    h#
  2360    3720    sh
  3720    5200    iy
  5200    6160    hv
  6160    8720    ae
  8720    9680    dcl
  9680   10173    y
 10173   11077    axr
 11077   12019    dcl
 12019   12257    d
```

A section of the corresponding annotation graph appears in Figure 3. Each node displays the node identifier and the time offset. The arcs are decorated with type and label information. Type W is for words and the type P is for phonetic transcriptions.

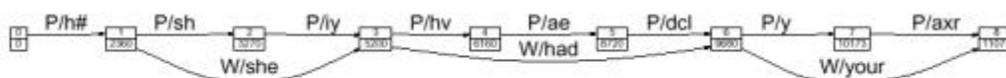

**Figure 3: A TIMIT annotation graph**

Since an annotation graph is just a set of (timed) nodes, arcs and labels, it can be trivially represented using three relational tables:

```
Time:           Arc:                Label:
 N      T        A    X    Y   T      A   L
--------        --------------      -------
 0      0        1    0    1   P      1   h#
 1    2360       2    1    2   P      2   sh
 2    3270       3    2    3   P      3   iy
 3    5200       4    3    4   P      4   hv
 4    6160       5    4    5   P      5   ae
 5    8720       6    5    6   P      6   dcl
 6    9680       7    6    7   P      7   y
 7   10173       8    7    8   P      8   axr
 8   11077       9    8    9   P      9   dcl
 9   12019      10    9   10   P     10   d

10   12257      19    3    6   W     18   she
14   16626      20    6    8   W     19   had
17   22179      21    8   14   W     20   your
                22   14   17   W     21   dark
                                     22   suit
```

A large amount of annotation can be efficiently represented and indexed in this manner. This brings us to the question of converting (or loading) existing data into such a database. The LDC's catalog alone includes nearly 200 publications, where each typically has its own format (often more than one). The sheer quantity and diversity of the data presents a significant challenge to the conversion process. In addition, some corpora exist in multiple versions, or include uncorrected, corrected and re-corrected parts.

The Annotation Graph Toolkit, version 1.0, contains a complete implementation of the annotation graph model, import filters for several formats, loading/storing data to an annotation server (MySQL), application programming interfaces in C++ and Tcl/tk, and example annotation tools for dialogue, ethology and interlinear text. The supported formats are: xlabel, TIMIT, BAS Partitur, Penn Treebank, Switchboard, LDC Callhome, CSV and AIF level 0. Future work will provide Python and Perl interfaces, more supported formats, a query language and interpreter, and a multi-channel transcription tool. All software is distributed under an open source license, and is available from http://www.ldc.upenn.edu/AG/.

Given that the annotation data can be stored in a relational database, it can be queried directly in SQL. More convenient, a domain-specific query language will be developed (see Cassidy and Bird 2000 and the work cited there). Query expressions will be transmitted over the web in the form of a CGI request, and translated into SQL by the annotation server. The resulting annotation data will be returned in the form of an XML document. An example for the TIMIT database, using the language proposed by Cassidy and Bird (2000), will serve to illustrate:

*Find word arcs spanning a sequence of segments beginning with hv and containing ae:*

```
http://BASE-URL/cgi-bin/query?
X.[].Y<timit/word;
X.[:hv].[]*.[:ae].[]*.Y<-timit/ph
```

Executed on the above annotation data, this query would return the XML document in Figure 4.

Neither the query nor the returned document are intended for human consumption. A client-side annotation tool will initiate queries and display annotation content on behalf of an end-user.

```
<?xml version="1.0"?>
<!DOCTYPE AGSet SYSTEM "ag.dtd">
<AGSet id="Timit" version="1.0" xmlns="http://www.ldc.upenn.edu/atlas/ag/"
       xmlns:xlink="http://www.w3.org/1999/xlink"
       xmlns:dc="http://purl.org/DC/documents/rec-dces-19990702.htm">
<Timeline id="T1">
<Signal id="S1" mimeClass="audio" mimeType="wav" encoding="wav"
        unit="16kHz" xlink:href="TIMIT/train/dr1/fjsp0/sa1.wav"/>
</Timeline>
<AG id="t1" type="transcription" timeline="T1">
<Anchor id="A3" offset="5200" unit="16kHz"/>
<Anchor id="A6" offset="9680" unit="16kHz"/>
<Annotation id="Ann10" type="W" start="A3" end="A6">
<Feature name="label">had</Feature>
</Annotation>
</AG>
</AGSet>
```

**Figure 4: Document returned by AG query**

This annotation tool and server are integrated using the model shown below. A simplified client-server model, working at the level of annotation files is already available with the current distribution of the Annotation Graph Toolkit. Significantly, a networked annotation tool is identical to a standalone version, except that the AG library fetches its data from a remote server instead of local disk.

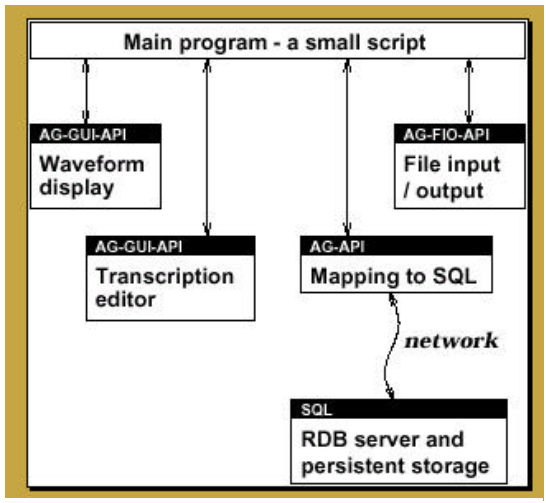

**Figure 5: Interactions among annotation tools and the annotation server**

The annotation graph formalism, annotation servers and the emerging query language will provide basic infrastructure to store, process and deliver essentially arbitrary amounts and types of signal annotations for a wide variety of research and teaching tasks including those described above. This infrastructure will enable reuse of existing resources and coordinated development of new resources both within and across research communities working with annotated linguistic datasets.

# 5  Remaining Challenges to Language Resource Development

We have described a process whereby annotated data in a variety of formats can be loaded into a central database server that interacts directly with annotation tools. The Annotation Graph Toolkit, version 1.0, is the first implementation of this architecture. As the toolkit undergoes future development, it will need to deal continually with conversion issues.

Annotation data will continue to be created and manipulated by multiple tools and to be stored in incompatible file formats. Data will continue to be mapped between different formats so that appropriate tools can be used, and appropriately managed to keep inconsistencies from arising. There will still be times when we need to trace the provenance of a particular item, back through a history involving several formats. These will always be hard problems; the proposed infrastructure will address them but no infrastructure is likely to eliminate conversion, integrity and provenance issues.

Annotation graphs focus on the problems of dealing with time series. They do not directly address paradigmatic data such as lexicons and demographic tables. One should note however, that time series data and paradigmatic data can be united efficiently. As already mentioned, annotation graphs may be stored trivially in relational tables, technology routinely used for paradigmatic data. In this way, conventional "joins" of relational table can convolve time-series annotations with paradigms (e.g. texts with dictionaries or utterances with speaker demographics).

Through judicious compromises - such as one-time computer-assisted conversion of legacy annotation data and creating once-off interfaces to existing useful tools - and through the judicious combination of simple and well-supported formalisms and technologies as described above, we believe that the management problems can be substantially reduced in scale and severity.

We can illustrate the advantages of AG with a example of the annotation of the Switchboard corpus for –t/d deletion. Switchboard contains two-channel audio of thousands of 5-minute conversations among pairs of speakers that have been transcribed with the transcripts time-aligned to the audio. A single utterance is written:

```
274.35  279.50  A.119  Uh, he,
uh, carves out different figures
in the, in the plants,
```

giving the start and stop time of the utterance, channel, speaker ID and the transcript of the utterance. This can be converted trivially into AG format as above.

The DASL tool concordances audio transcripts and identifies utterances in which the target phenomenon (eg. –t/d deletion) may occur. A line of the concordance file contains two IDs one to identify the utterance within the concordance, the other to link back to the original corpus. The <annotate> tags identify a potential environment for the phenomenon under study.

```
<sample id="1" senid="10194">uh
he uh carves out <annotate>
different figures </annotate> in
the in the p[lants]- plants
shrubs </sample>
```

The link between the concordance and the original corpus is maintained through a table containing: Sentence_ID, File_ID, Start_Time, Stop_Time, Channel and Speaker.

```
10194 2141 274.35  279.50 A 1139
```

Speakers' demographic data appears in another table containing: Speaker_ID, Sex, Age, Region, Education_Level

```
1139, MALE, 50, NORTHERN, 2
```

The DASL interface embeds the concordance results in a template containing input fields for each parameter to be annotated (see Figure 2). The linguist's annotation of the utterance can be stored in AG formalism as in Figure 5. Note that although AGs provide an elegant and general solution to the annotation of time series data, they do not remove the need to deal with the ad hoc formats one may encounter in various corpora. Nor do they remove the need to track the relations among elements in time-series data and paradigmatic material.

## 6 Conclusions

Researchers in human language share assumptions and needs within and across research communities. Each group feels an acute need for language resources including data, annotations, formats and processes. This paper has summarized some common needs and described an architecture for encoding annotations and delivering them via annotation servers using SQL or a custom query language. Much of the architecture discussed has already been created and is available in the Annotation Graphic Toolkit. Other components, especially the query language, are currently under development. It is hoped that tools based on annotations graphs and annotation servers will encourage greater levels of resource sharing and the coordination of future resource development.

```
<?xml version="1.0"?>
<!DOCTYPE AGSet SYSTEM "http://www.ldc.upenn.edu/AG/dom/xml/ag.dtd">
<AGSet id="DASL" version="1.0"
       xmlns="http://www.ldc.upenn.edu/atlas/ag/" xmlns:xlink="http://www.w3.org/1999/xlink"
       xmlns:dc="http://purl.org/DC/documents/rec-dces-19990702.htm">
<Metadata></Metadata>
<Timeline id="DASL:Timeline1"> <Signal id="DASL:Timeline1:Signal1" mimeClass="audio"
       mimeType="wav" encoding="mu-law" unit="8kHz" xlink:type="simple"
       xlink:href="LDC93S7:sw2141.wav">
</Signal></Timeline>
<AG id="DASL:AG1" timeline="DASL:Timeline1">
<Anchor id="DASL:AG1:Anchor1" offset="274.595" signals="DASL:Timeline1:Signal1"></Anchor>
<Anchor id="DASL:AG1:Anchor2" offset="280.671" signals="DASL:Timeline1:Signal1"></Anchor>
<Annotation id="DASL:AG1:Annotation1" type="csv" start="DASL:AG1:Anchor1"
       end="DASL:AG1:Anchor2">
<Feature name="td">Deleted</Feature>          <Feature name="Morphological">Monomorpheme</Feature>
<Feature name="EPreceding">AlveolarNasal</Feature> <Feature name="EFollowing">Obstruent</Feature>
<Feature name="Same_Prec_Foll">N/A</Feature>      <Feature name="Stress">Unstressed</Feature>
<Feature name="Cluster_complexity">Two_elements</Feature>
<Feature name="Sentence_id">1</Feature>           <Feature name="Corpus_name">swb</Feature>
<Feature name="WPreceding">uh he uh carves out </Feature>
<Feature name="WMatched">different figures</Feature>
<Feature name="WFollowing"> in the in the p[lants]- plants shrubs</Feature>
<Feature name="File_name">/speech/swb0/sw2141.wav</Feature>
<Feature name="Speech_channel">1</Feature>        <Feature name="Speaker_id">1139</Feature>
<Feature name="Sex">MALE</Feature>              <Feature name="Birth_year">1956</Feature>
<Feature name="Dialect">NORTHERN</Feature>      <Feature name="Edu">2</Feature>
</Annotation></AG></AGSet>
```

**Figure 5: A sociolinguistic annotation in AG format**